\documentclass[runningheads]{llncs}

\usepackage{eccv}

\usepackage{eccvabbrv}

\usepackage{graphicx}
\usepackage{booktabs}

\usepackage[accsupp]{axessibility}  

\usepackage{hyperref}

\usepackage{orcidlink}

\graphicspath{{figures/}} 
\usepackage{multirow} 
\usepackage{adjustbox}

\begin{document}

\title{NimbleD: Enhancing Self-supervised Monocular Depth Estimation with Pseudo-labels and Large-scale Video Pre-training} 

\titlerunning{NimbleD}

\author{Albert Luginov \orcidlink{0009-0002-7243-9808} \and
Muhammad Shahzad \orcidlink{0009-0002-9394-343X}}

\authorrunning{A. Luginov and M. Shahzad}

\institute{
University of Reading\\
\email{a.luginov@pgr.reading.ac.uk}\\
\email{m.shahzad2@reading.ac.uk}
}

\maketitle

\begin{abstract}
We introduce NimbleD, an efficient self-supervised monocular depth estimation learning framework that incorporates supervision from pseudo-labels generated by a large vision model. This framework does not require camera intrinsics, enabling large-scale pre-training on publicly available videos. Our straightforward yet effective learning strategy significantly enhances the performance of fast and lightweight models without introducing any overhead, allowing them to achieve performance comparable to state-of-the-art self-supervised monocular depth estimation models. This advancement is particularly beneficial for virtual and augmented reality applications requiring low latency inference. The source code, model weights, and acknowledgments are available at \url{https://github.com/xapaxca/nimbled}.

\end{abstract}

\section{Introduction}
\label{sec:intro}

Monocular depth estimation (MDE) is the task of predicting the distances to objects relative to the camera from a single image input. Low latency depth estimation is crucial for metaverse applications as it ensures real-time, accurate spatial awareness and immersive interactions with virtual and real-world objects, enhancing user experience and comfort. Recent advancements in vision transformers (ViT)~\cite{vit}, large vision models~\cite{clip}, and generative models~\cite{stable_diffusion} have significantly advanced MDE~\cite{depthanything, midas31, marigold, zerodepth, zoedepth, unidepth, dmd, 3rd_mdec}. The two main training paradigms commonly used for MDE are supervised and self-supervised learning (SSL). Supervised MDE relies on ground truth depth labels obtained from LiDAR or RGB-D cameras, whereas SSL MDE leverages geometric constraints from monocular videos or stereo setups. On one hand, SSL MDE models~\cite{sfmlearner} trained exclusively on monocular videos represent the most accessible approach since they require only monocular videos for training, without the need for a stereo setup and ground truth depth, potentially allowing for training on large-scale unlabeled data~\cite{slowtv}. On the other hand, the recent supervised models employing large-scale training~\cite{depthanything} or generative methods~\cite{marigold} have demonstrated superior zero-shot depth estimation capabilities. However, these models are often relatively slow at inference, limiting their use in metaverse applications where low latency is crucial. Another challenge is the feasibility of training these models for companies or individuals with limited resources. In this work, we address these challenges by introducing NimbleD, a straightforward yet effective SSL MDE learning framework enhanced by pseudo-labels generated by a large vision model and large-scale video pre-training. This framework noticeably boosts the depth estimation quality of fast and lightweight models without introducing any overhead, enabling them to achieve the performance of larger state-of-the-art SSL MDE methods.

Our contributions are summarized as follows:
\begin{itemize}
\item Develop an effective training strategy that utilizes both self-supervision from monocular videos and pseudo-supervision from disparities generated by a large vision model.

\item Introduce a concise loss function that combines SSL and PSL losses.

\item Incorporate large-scale video pre-training from publicly available videos\\without any complex data preparation.

\item Boost the depth estimation performance of fast and lightweight models to a level comparable to state-of-the-art SSL MDE methods.
\end{itemize}

\section{Related Work}

Recent advancements in zero-shot MDE have yielded significant progress. MiDaS~\cite{midas, midas31} has effectively addressed training challenges on diverse datasets by implementing a scale- and shift-invariant loss, ensuring compatibility with various ground truth forms. Similarly, ZoeDepth's~\cite{zoedepth} innovative two-stage framework integrates relative and metric depth estimation, utilizing large-scale pre-training alongside domain-specific fine-tuning, augmented by a novel bin adjustment mechanism for scale-aware predictions. ZeroDepth's~\cite{zerodepth} introduction of input-level geometric embeddings and a variational latent representation has facilitated scale-aware depth estimation across diverse domains, leveraging extensive labeled and synthetic datasets. UniDepth~\cite{unidepth} introduces a module that learns camera parameters and uses a pseudo-spherical representation to separate the camera and depth components, demonstrating superior metric depth estimation. KBR's~\cite{slowtv} self-supervised learning approach capitalizes on a vast dataset derived from YouTube, covering a wide array of environments without necessitating camera intrinsics. Marigold~\cite{marigold} employs diffusion-based image generators alongside a fine-tuned Stable Diffusion~\cite{stable_diffusion} model with synthetic data, whereas DMD's~\cite{dmd} diffusion-based method incorporates field-of-view augmentation and log-scale depth representation, further refined through fine-tuning on a varied dataset mix. DepthAnything~\cite{depthanything} stands out by proposing a large-scale training regimen for DPT~\cite{dpt}, utilizing an extensive dataset coupled with knowledge distillation, semantic supervision, and CutMix~\cite{cutmix} data augmentation, showcasing a wide-ranging and innovative approach within the field.

Self-supervised MDE trained exclusively on monocular videos was first introduced by Zhou~\etal~\cite{sfmlearner}. Monodepth2~\cite{monodepth2} further advanced the field by addressing challenges associated with occluded objects and objects moving at the same speed as the camera. Following these foundational works, significant progress has been made. This includes recent works focusing on depth network architecture enhancements~\cite{sqldepth, daccn} and incorporation of additional supervisory signals~\cite{gasmono, FGTO}.

There are studies aimed at developing lightweight MDE models that employ mobile architectures. LiteMono~\cite{litemono} utilizes a custom hybrid CNN-Transformer encoder, while SwiftDepth~\cite{swiftdepth} uses SwiftFormer~\cite{swiftformer} as an encoder accompanied by a custom two-stage decoder. Both models exhibit competitive depth estimation quality with fewer parameters and faster inference speeds, making them ideal for augmented and virtual reality applications.

Various works explore knowledge distillation techniques. Wang \etal~\cite{wang_kd} propose a method that distills knowledge from the teacher model at both the feature and output levels, utilizing silog~\cite{eigen_metrics}, gradient matching~\cite{midas}, and pairwise distillation~\cite{pairwise_distillation_loss} losses. TIE-KD~\cite{tie_kd} focuses on output-level knowledge distillation, employing Kullback-Leibler divergence and SSIM~\cite{ssim} for the loss function. MViTDepth~\cite{mvitdepth} combines self-supervised learning with knowledge distillation, using the logarithm of the L1 difference at the output level. MAL~\cite{mal} is a multi-frame MDE method that leverages additional supervision from a instance segmentation model to identify moving objects and performs knowledge distillation at the output level using the L1 difference.

In this work, we address the progress in zero-shot MDE by leveraging a modern large vision model to generate pseudo-labels, introducing a straightforward yet efficient loss function, and applying large-scale video pre-training. Our approach ensures that lightweight self-supervised MDE methods match the depth estimation quality of larger state-of-the-art models, making them suitable for real-time applications in the metaverse.

\section{Method}

\begin{figure}[t]
\centering 
\includegraphics[width=\linewidth]{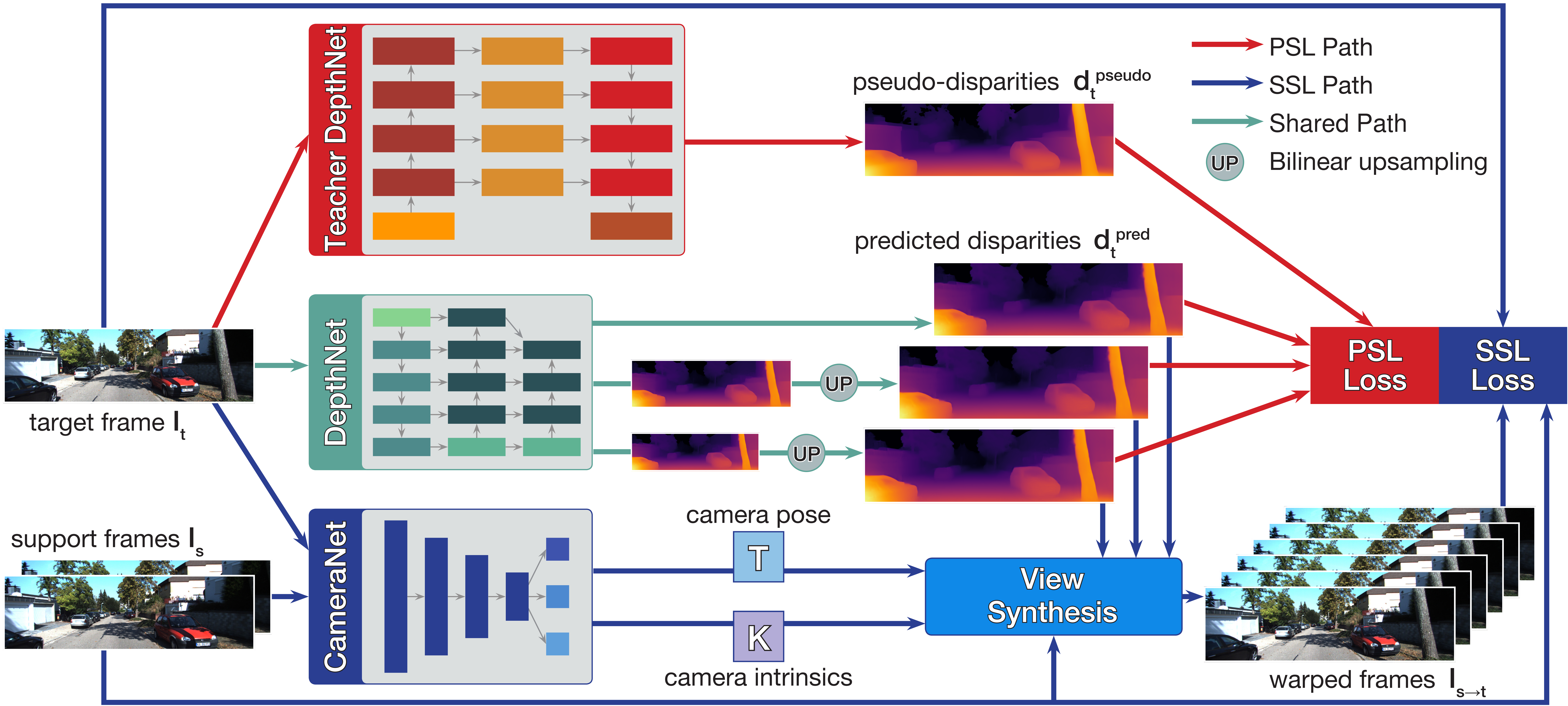}
\caption{Overview of the proposed NimbleD learning framework. The framework comprises a depth network that outputs multi-scale disparity predictions ($d_t^{\text{pred}}$), a camera network that outputs the relative camera pose ($T$) and camera intrinsic parameters ($K$), and a teacher depth network that generates pseudo-disparities ($d_t^{\text{pseudo}}$). The ultimate objective of the method is to minimize the loss between the predicted and pseudo-disparities ($\textit{PSL Loss}$) and to minimize the image reconstruction loss ($\textit{SSL Loss}$).}
\label{fig:framework}
\end{figure}

In this section, we introduce NimbleD learning framework. First, the problem statement of self-supervised MDE enhanced by pseudo-labels is introduced. Next, the combined SLL and PSL loss function is detailed. Finally, the large-scale video pre-training without camera intrinsics is described.

\subsection{Self-superivsed Monocular Depth Estimation Enhanced by Pseudo-labels} 

Self-supervised MDE is generally formulated as an image reconstruction problem. Depth estimation serves as a building block of the entire training framework, which is then used independently during inference. The framework consists of a depth network that predicts disparities \(d_t\) from a target frame \(I_t\) and a pose network that predicts the relative camera pose \(T\) between the target frame \(I_t\) and support frames \(I_s\), where \(s \in \{t-1, t+1\}\) indicates the frames temporally adjacent to the target frame. Given depth map \(D_t\), which is the inverse of disparities (\(1 / d_t\)), camera pose \(T\), and camera intrinsics \(K\) that must usually be known beforehand, the target frame is reconstructed from support frames \(I_s\) as follows:
\begin{equation}
I_{s\rightarrow t} = I_s \langle \text{proj}(D_t, T, K) \rangle,
\end{equation}
where \(I_{s\rightarrow t}\) represents the reconstructed target image, \(\text{proj}()\)
denotes the projection of depth onto the support frames \(I_s\) coordinates, and \(\langle\rangle\) signifies a differentiable sampling operator~\cite{stn}.

Following~\cite{slowtv}, the pose network is assigned to learn camera intrinsics \(K\). The only modification introduced to the network involves adding two additional output heads to the decoder: the first outputs the focal lengths, and the second outputs the principal point coordinates, which together construct the intrinsics matrix \(K\). Since it predicts all camera parameters (both extrinsic and intrinsic), we refer to it as the camera network.

Supervised MDE is generally formulated as the minimization of error between predicted and ground truth depth maps. Considering situations where ground truth information is unavailable, we employ a large vision model to generate pseudo-labels and minimize the error between predictions and these pseudo-labels. This pseudo-supervisory signal enhances our self-supervised image reconstruction problem and can be seen as offline knowledge distillation at the output level.

The proposed NimbleD learning framework is depicted in~\cref{fig:framework}.

\subsection{Combined Self-superivsed and Pseudo-supervised Loss Function}

We propose a simple yet effective combination of losses widely used in self-supervised and supervised MDE. Our total loss is a weighted average of SSL and PSL losses, averaged across the upsampled multi-scale outputs of the depth network. It is defined as:  

\begin{equation}
L_{\text{total}} = \frac{1}{n} \sum_{i=1}^{n} (\lambda L_{\text{SSL}}^i + (1 - \lambda) L_{\text{PSL}}^i),
\end{equation}
where $\lambda$ is the loss weighting parameter, and $n$ denotes the number of output scales.

The main building block of the SSL loss is photometric error, defined as:
\begin{equation}
\begin{aligned}
\text{pe}(I_t, I_{s\rightarrow t}) = \frac{\alpha}{2} \left(1 - \text{SSIM}(I_t, I_{s\rightarrow t})\right) + (1 - \alpha) ||I_t - I_{s\rightarrow t}||_1,
\end{aligned}
\end{equation}
where $\text{SSIM}$ denotes the Structural Similarity Index Measure~\cite{ssim}, and $\alpha$ is set to $0.85$.

The SSL loss is the per-pixel minimum reprojection loss with\\auto-masking~\cite{monodepth2}, defined as:
\begin{equation}
L_{\text{SSL}} = \text{mean} \left( \mu \min_{s \in \{t-1, t+1\}} \text{pe}(I_t, I_{s \rightarrow t}) \right),
\end{equation}
where $\mu$ represents the auto-masking function, given by:
\begin{equation}
\mu = [\min_{s\in\{t-1, t+1\}} \text{pe} (I_t, I_{s\rightarrow t}) < \min_{s\in\{t-1, t+1\}} \text{pe} (I_t, I_{s})],
\end{equation}
where $[\cdot]$ is the indicator function, which equals 1 if the enclosed condition is true and 0 otherwise.

The PSL loss is defined as a scale-and-shift invariant loss~\cite{midas}, calculated in the disparity space:
\begin{equation}
L_{\text{PSL}} = \text{mean} \left( \left| \hat{d}_{\text{pred}} - \hat{d}_{\text{pseudo}} \right| \right),
\end{equation}
where $\hat{d}_{\text{pred}}$ and $\hat{d}_{\text{pseudo}}$ are the scaled and shifted predicted and\\pseudo-disparities, respectively. These are defined as follows:
\begin{equation}
\hat{d} = \frac{d - \text{median}(d)}{\text{mean} \left( \left| d - \text{median}(d) \right| \right)}.
\end{equation}

It is important to note that widely used smoothness regularization loss~\cite{monodepth} was not employed due to its negligible impact on the results~\cite{spencer_benchmark}, see ablation studies. This decision is made to save training time during the large-scale video pre-training.

\subsection{Large-Scale Video Pre-training}

\begin{figure}[t]
\centering 
\includegraphics[width=\columnwidth]{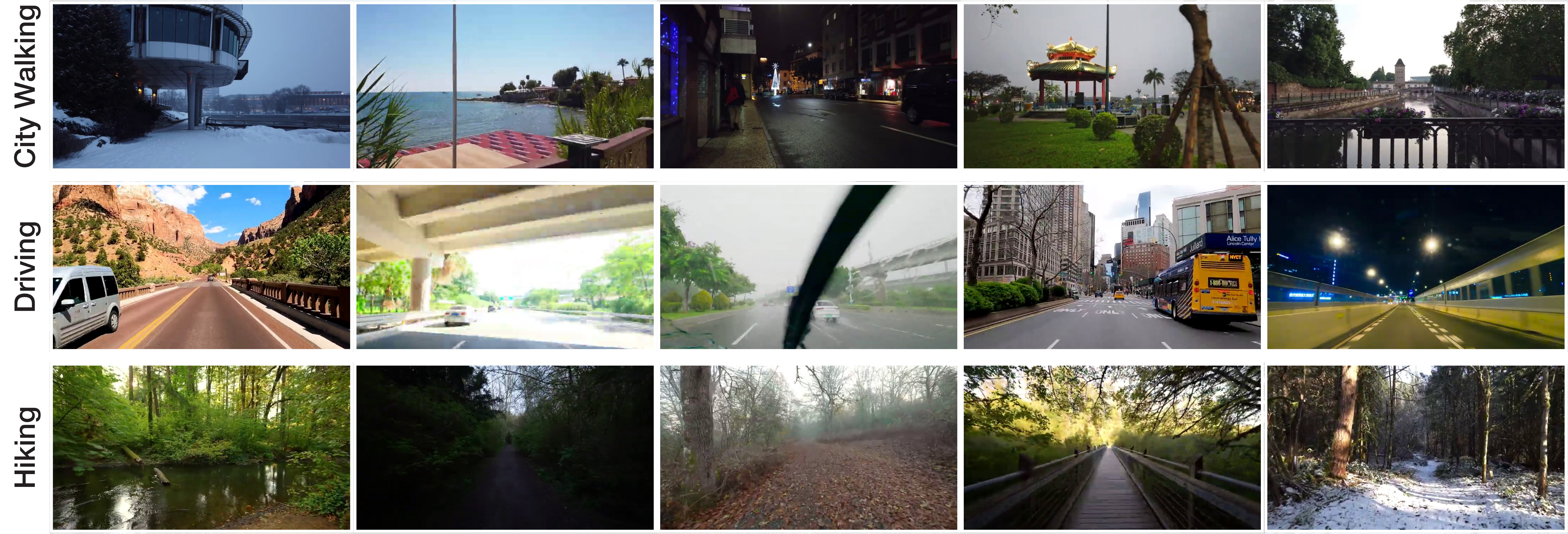}
\caption{The dataset for large-scale video pre-training is curated from publicly available videos. It consists of three equally represented classes - city walking, driving, and hiking - offering a diverse range of outdoor environments.}
\label{fig:dataset}
\end{figure}

\begin{table}[t]
\centering
\caption{Summary of the dataset for large-scale video pre-training. The dataset includes three balanced classes, each characterized by a diverse set of filming conditions, totaling more than 1.1 million frame triplets.}
\begin{tabular}{lccccc}
    \toprule
    \multirow{2.5}{*}{Class} & \multirow{2.5}{*}{\#Videos} & \multirow{2.5}{*}{Duration (h)} & \multirow{2.5}{*}{\#Frames} & \multicolumn{2}{c}{\multirow{1}{*}{\#Triplets}} \\
    \cmidrule(lr){5-6}
     & & & & Large set & Small set \\
    \midrule
    City walking & 35 & 23.1 & 383,199 & 363,973 & 19,156 \\
    Driving & 35 & 23.1 & 400,617 & 380,518 & 20,029 \\
    Hiking & 35 & 23.1 & 392,446 & 372,758 & 19,618 \\
    Total & 105 & 69.7 & 1,176,262 & 1,117,249 & 58,803 \\
    \bottomrule
\end{tabular}
\label{tab:dataset}
\end{table}

NimbleD's camera network, which predicts camera intrinsics, enables the use of any video with moving cameras and objects. Therefore, we select YouTube videos with a CC-BY license for large-scale pre-training.

We categorize the selected videos into three classes to ensure a diverse range of outdoor environments: driving, hiking, and city walking. The driving class comprises videos from the USA and China, featuring a variety of driving conditions. The hiking videos, filmed in the woods of national parks across the USA, provide natural, uneven terrain. The city walking videos, covering urban areas in Europe and Asia, offer a mix of architectural styles and pedestrian scenarios. Each class includes 35 videos, with each totaling approximately 23 hours, resulting in a 70-hour dataset. Unlike SlowTV~\cite{slowtv}, our dataset ensures a more balanced composition by maintaining a similar number of frames in each class, as summarized in~\cref{tab:dataset}.

The videos are downloaded at a resolution of 480$\times$854 to manage space efficiently, and frames are extracted at a rate of 5 fps to balance motion detail with dataset size. The dataset encompasses a variety of lighting conditions, weather scenarios, and filming challenges. For example, some driving videos are filmed with a camera installed inside a vehicle, capturing reflections and raindrops on the windshield, as well as the movement of the wipers. City walking videos may display abrupt camera movements, resulting in reduced overlap between frames. Hiking footage often suffers from motion blur and defocusing issues.

Despite these challenges, we decide against pre-processing the videos or the extracted frames, aiming to present the data in its most authentic form, embracing the imperfections and variability inherent in real-world footage. This approach ensures that the dataset reflects genuine outdoor conditions, making it a valuable resource for developing and testing self-supervised MDE models. Some samples from the dataset are presented in~\cref{fig:dataset}.

Our dataset ultimately comprises 1.1M overlapping triplets. Additionally, we create a smaller set of 59K triplets for debugging.

\section{Experiments}

In this section, we first briefly introduce the datasets used and the evaluation metrics. Then, we describe the training implementation details, followed by the evaluation results. Finally, an extensive ablation study is performed to explore various components of our framework.

\subsection{Datasets}

In addition to the aforementioned selected publicly available videos, we utilize KITTI~\cite{kitti}, NYUv2~\cite{nyuv2}, and Make3D~\cite{make3d}.

For training on KITTI, we use Eigen-Zhou split~\cite{sfmlearner}, which consists of 39,810 automotive outdoor frame triplets. For evaluation on KITTI, we employ Eigen split~\cite{eigen_split}, which includes 697 frames with LiDAR-based ground truth depth, adhering to the standard evaluation protocol. This protocol includes Garg crop~\cite{garg}, 80-meter distance cap, and input resolution of 192×640. Additionally, for comparison with the teacher model, we use Eigen-Benchmark split~\cite{eigen_benchmark_split}, which comprises 652 frames with improved LiDAR-based ground truth depth, and in this context, we forego cropping.

For zero-shot generalization, we turn to NYUv2 and Make3D, which are not seen by the teacher depth network. NYUv2 encompasses 654 indoor frames with ground truth depth sourced from an RGB-D camera. This evaluation leverages Eigen crop~\cite{eigen_metrics}, 10-meter distance cap, and input resolution of 288×384. Make3D consists of 134 outdoor frames with laser-based ground truth depth. We use 70-meter distance cap, center crop, and input resolution of 192×640.

Across these evaluation scenarios, we consistently employ a set of standard evaluation metrics~\cite{eigen_metrics}. For NYUv2 and Make3D, we use AbsRel, SqRel, RMS, and RMS$_{\log}$. For evaluation on KITTI, we additionally use $\delta_1<1.25$, $\delta_2<1.25^2$, and $\delta_3<1.25^3$.

\subsection{Implementation Details}

\begin{table}[t]
\centering
\caption{Training details. Training consists of two stages: pre-training on publicly available YouTube videos and fine-tuning on KITTI~\cite{kitti}. $LR$ stands for learning rate.}
\begin{tabular}{lcc}
    \toprule
    ~ & Pre-training & Fine-tuning  \\
    \midrule
    Dataset & YouTube videos & KITTI~\cite{kitti} \\
    Color Jitter & $\checkmark$ & $\checkmark$ \\
    Horizontal flip & $\checkmark$ & $\checkmark$ \\
    Aspect augmentation & $\checkmark$ & ~ \\
    Batch size & 16 & 16 \\
    \#Epochs & 6 & 20 \\
    Training time (h) & 32 & 5 \\
    Optimizer & AdamW\cite{adamw} & AdamW\cite{adamw} \\
    Weight decay & $10^{-3}$ & $10^{-3}$ \\
    Initial LR & $10^{-4}$ & $10^{-4}$ \\
    LR scheduler & Cosine\cite{cosine} & Step \\
    Adjusted LR & $10^{-5}$ & $10^{-5}$ \\
    Adjustment epoch & 4 & 10 \\
    Loss weighting $\lambda$ & 0.9 & 0.9 (0.95 for last 5 ep.) \\
    \bottomrule
\end{tabular}
\label{tab:training_parameters}
\end{table}

For the teacher depth network, we utilize DepthAnything~\cite{depthanything}, chosen for its high-quality results. Specifically, its ability to generate a large volume of pseudo-labels with limited computational resources stands out, contrasting with diffusion-based models~\cite{marigold}. Notably, DepthAnything is not trained on the KITTI~\cite{kitti}, NYUv2~\cite{nyuv2}, or Make3D~\cite{make3d} datasets. We observe its optimal performance at a resolution of 518$\times$518, rather than at the original input image size. Therefore, we first resize the original images to 518$\times$518 and then interpolate the predictions back to the original image size. To maintain the integrity of the pseudo-labels, we store the disparities without normalization and also preserve the maximum value to facilitate information retrieval.

For the depth network, we utilize several baseline methods: Monodepth2~\cite{monodepth2}, SwiftDepth~\cite{swiftdepth}, and LiteMono~\cite{litemono}, each with different configurations. Monodepth2 is the standard baseline model for SSL MDE research, while SwiftDepth and LiteMono are fast and lightweight models. The encoders of all depth networks are pre-trained on ImageNet~\cite{imagenet}. To ensure that the predicted disparities are consistent with the teacher model, we replace the output activation function from sigmoid to ReLU, thereby eliminating the need for output rescaling to a specific range.

For the camera network, we employ the widely-used PoseNet~\cite{sfmlearner, monodepth2}, based on an ImageNet~\cite{imagenet} pre-trained ResNet-18~\cite{resnet} backbone, to ensure compatibility with established methods. We modify the decoder to also output camera intrinsics, in line with KBR~\cite{slowtv}.

For training, we apply standard data augmentations~\cite{monodepth2} to the frames fed into the depth network. Additionally, for large-scale video pre-training, we employ aspect ratio augmentations similar to those used in KBR~\cite{slowtv}, involving center cropping at randomly selected aspect ratios followed by interpolation to sizes that approximate the total pixel counts of a 192$\times$640 input size.

Training is conducted on a single NVIDIA GeForce RTX 4090 GPU. All training details are summarized in~\cref{tab:training_parameters}.

\subsection{Results} 

\begin{table}[t]
\centering
\caption{Results on KITTI~\cite{kitti} Eigen split~\cite{eigen_split} with median alignment. The input image size is $192\times640$. $Par$ indicates the total number of parameters in millions. The highest scores are highlighted in \textbf{bold}, while the second-highest scores \underline{underlined}.}
\begin{tabular}{lccccccccc}
    \toprule
    \multirow{2.5}{*}{Method} & \multirow{2.5}{*}{Venue} & \multicolumn{5}{c}{Lower is better} & \multicolumn{3}{c}{Higher is better} \\
    \cmidrule(lr){3-7} \cmidrule(lr){8-10}
     &  & Par & AbsRel & SqRel & RMS & RMS$_{\log}$ & $\delta_1$ & $\delta_2$ & $\delta_3$ \\
    \midrule
    DaCCN~\cite{daccn} & ICCV`23 & 13.0 & 0.099 & \underline{0.661} & 4.316 & 0.173 & 0.897 & 0.967 & \underline{0.985} \\ 
    GasMono~\cite{gasmono} & ICCV`23 & 27.9 & 0.098 & - & 4.303 & 0.173 & 0.903 & 0.968 & 0.984 \\ 
    FGTO~\cite{FGTO} & CVPR`24 & 27.9 & 0.096 & 0.696 & 4.327 & 0.174 & 0.904 & 0.968 & \underline{0.985} \\ 
    SQLdepth~\cite{sqldepth} & AAAI`24 & 31.1 & \textbf{0.091} & 0.713 & \underline{4.204} & \underline{0.169} & \textbf{0.914} & 0.968 & 0.984 \\ 
    \midrule
    MD2-R18~\cite{monodepth2} & ICCV`19 & 14.8 & 0.115 & 0.903 & 4.863 & 0.193 & 0.877 & 0.959 & 0.981 \\ 
    \textit{$+$NimbleD (ours)} & - & 14.8 & 0.100 & 0.739 & 4.440 & 0.175 & 0.898 & 0.967 & \underline{0.985} \\ 
    \midrule
    MD2-R50~\cite{monodepth2} & ICCV`19 & 34.6 & 0.110 & 0.831 & 4.642 & 0.187 & 0.883 & 0.962 & 0.982 \\ 
    \textit{$+$NimbleD (ours)} & - & 34.6 & 0.097 & 0.721 & 4.377 & 0.172 & 0.904 & 0.968 & \underline{0.985} \\ 
    \midrule
    SwiftDepth-S~\cite{swiftdepth} & ISMAR`23 & 3.6 & 0.110 & 0.830 & 4.700 & 0.187 & 0.882 & 0.962 & 0.982 \\ 
    \textit{$+$NimbleD (ours) }& - & 3.6 & 0.098 & 0.733 & 4.401 & 0.174 & 0.901 & 0.968 & \underline{0.985} \\ 
    \midrule
    SwiftDepth~\cite{swiftdepth} & ISMAR`23 & 6.4 & 0.107 & 0.790 & 4.643 & 0.182 & 0.888 & 0.963 & 0.983 \\ 
    \textit{$+$NimbleD (ours) }& - & 6.4 & 0.096 & 0.697 & 4.333 & 0.171 & 0.905 & \underline{0.969} & \textbf{0.986} \\ 
    \midrule
    LiteMono-S~\cite{litemono} & CVPR`23 & \textbf{2.5} & 0.110 & 0.802 & 4.671 & 0.186 & 0.879 & 0.961 & 0.982 \\ 
    \textit{$+$NimbleD (ours)} & - & \textbf{2.5} & 0.099 & 0.709 & 4.370 & 0.172 & 0.898 & 0.967 & \textbf{0.986} \\ 
    \midrule
    LiteMono~\cite{litemono} & CVPR`23 & \underline{3.1} & 0.107 & 0.765 & 4.561 & 0.183 & 0.886 & 0.963 & 0.983 \\ 
    \textit{$+$NimbleD (ours)} & - & \underline{3.1} & 0.096 & 0.684 & 4.304 & 0.171 & 0.903 & \underline{0.969} & \textbf{0.986} \\ 
    \midrule
    LiteMono-8M~\cite{litemono} & CVPR`23 & 8.8 & 0.101 & 0.729 & 4.454 & 0.178 & 0.897 & 0.965 & 0.983 \\ 
    \textit{$+$NimbleD (ours)} & - & 8.8 & \underline{0.092} & \textbf{0.646} & \textbf{4.194} & \textbf{0.165} & \underline{0.910} & \textbf{0.970 }& \textbf{0.986} \\ 
    \bottomrule
\end{tabular}
\label{tab:kitti_eigen} 
\end{table}

\begin{table}[t]
\centering
\caption{Results on KITTI~\cite{kitti} KITTI Eigen-Benchmark~\cite{eigen_benchmark_split} splits with improved ground truth against the teacher model using least-squares alignment. The input image size is $350\times350$ and $518\times518$ for DepthAnything~\cite{depthanything} and $192\times640$ for methods enhanced by NimbleD. $Par$ indicates the total number of parameters in millions. The highest scores are highlighted in \textbf{bold}, while the second-highest scores \underline{underlined}.}
\begin{tabular}{lcccccccc}
    \toprule
    \multirow{2.5}{*}{Method} & \multicolumn{5}{c}{Lower is better} & \multicolumn{3}{c}{Higher is better} \\
    \cmidrule(lr){2-6} \cmidrule(lr){7-9}
     & Par & AbsRel & SqRel & RMS & RMS$_{\log}$ & $\delta_1$ & $\delta_2$ & $\delta_3$ \\
    \midrule
    DepthAnything~\cite{depthanything} 350$\times$350  & 335.3 & 0.071 & 0.432 & 3.504 & 0.122 & 0.947 & 0.989 & \underline{0.997} \\ 
    DepthAnything~\cite{depthanything} 518$\times$518 & 335.3 & \textbf{0.064} & \textbf{0.363} & \textbf{3.239} & \underline{0.111} & \textbf{0.957} & \textbf{0.991} & \underline{0.997} \\ 
    \midrule
    MD2-R18~\cite{monodepth2} \textit{$+$NimbleD} & 14.8 & 0.073 & 0.448 & 3.610 & 0.122 & 0.941 & 0.988 & 0.996 \\ 
    MD2-R50~\cite{monodepth2} \textit{$+$NimbleD} & 34.6 & 0.070 & 0.438 & 3.571 & 0.134 & 0.945 & 0.989 & \underline{0.997} \\ 
    SwiftDepth-S~\cite{swiftdepth} \textit{$+$NimbleD} & 3.6 & 0.072 & 0.441 & 3.585 & 0.120 & 0.943 & 0.989 & \underline{0.997} \\ 
    SwiftDepth~\cite{swiftdepth} \textit{$+$NimbleD} & 6.4 & 0.069 & 0.408 & 3.469 & 0.117 & 0.946 & \underline{0.990} & \underline{0.997} \\ 
    LiteMono-S~\cite{litemono} \textit{$+$NimbleD} & \textbf{2.5} & 0.072 & 0.437 & 3.584 & 0.124 & 0.940 & 0.988 & \underline{0.997} \\ 
    LiteMono~\cite{litemono} \textit{$+$NimbleD} & \underline{3.1} & 0.070 & 0.427 & 3.537 & 0.126 & 0.944 & 0.989 & \underline{0.997} \\ 
    LiteMono-8M~\cite{litemono} \textit{$+$NimbleD} & 8.8 & \underline{0.065} & \underline{0.374} & \underline{3.338} & \textbf{0.110} & \underline{0.951} & \textbf{0.991} & \textbf{0.998} \\ 
    \bottomrule
\end{tabular}
\label{tab:kitti_eigen_benchmark_lsqr}
\end{table}

\begin{figure}[t!]
\centering 
\includegraphics[width=\linewidth]{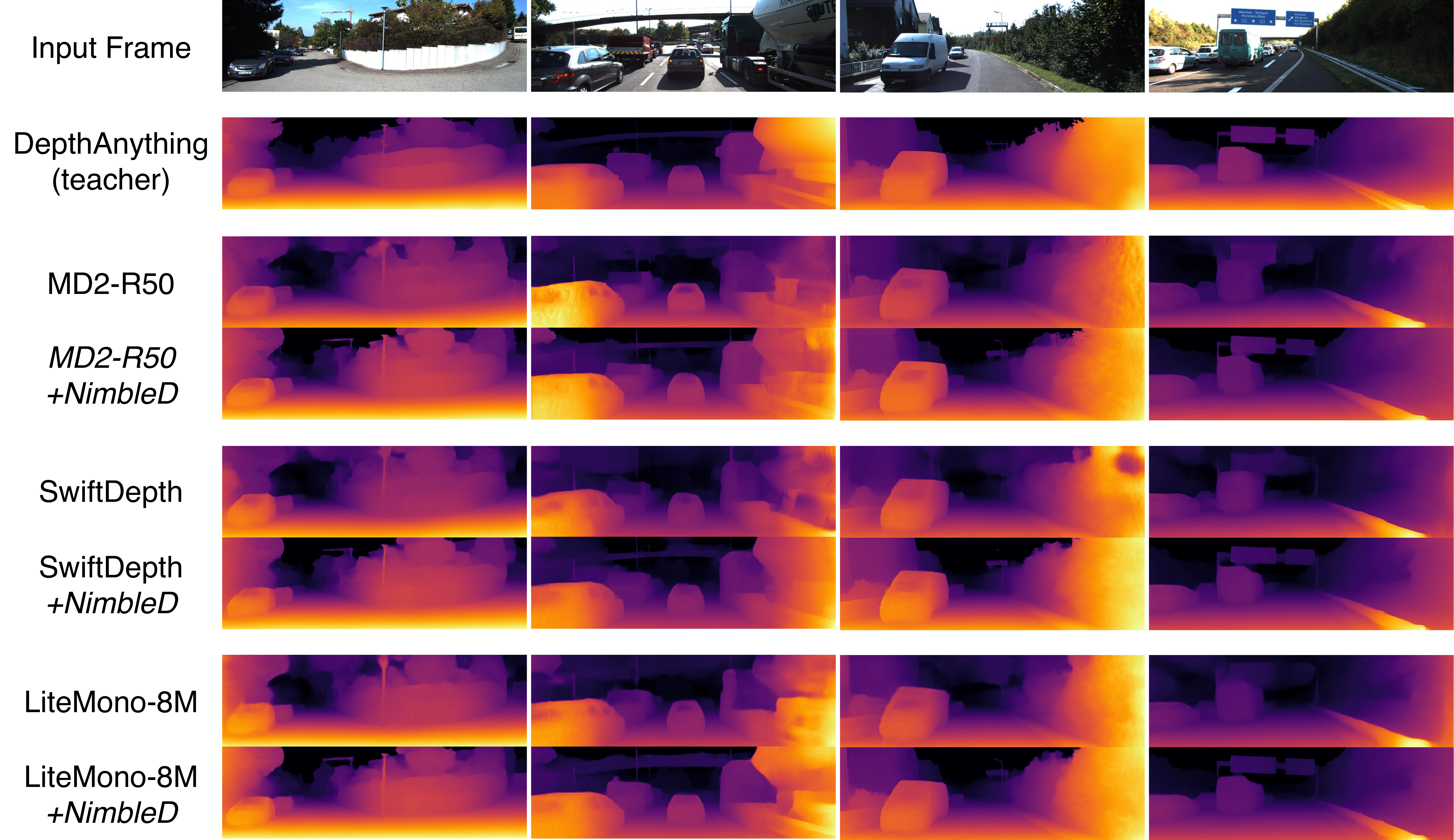}
\caption{Qualitative results on KITTI~\cite{kitti} Eigen split~\cite{eigen_split}, compared with the teacher model~\cite{depthanything} and baseline models~\cite{monodepth2, swiftdepth, litemono}. NimbleD observably enhances the depth estimation quality of all baseline models. It identifies distant objects not detected by the teacher model and demonstrates improved handling of sky regions compared to the baseline models.}
\label{fig:kitti}
\end{figure}

\begin{figure}[t]
\centering 
\includegraphics[width=\linewidth]{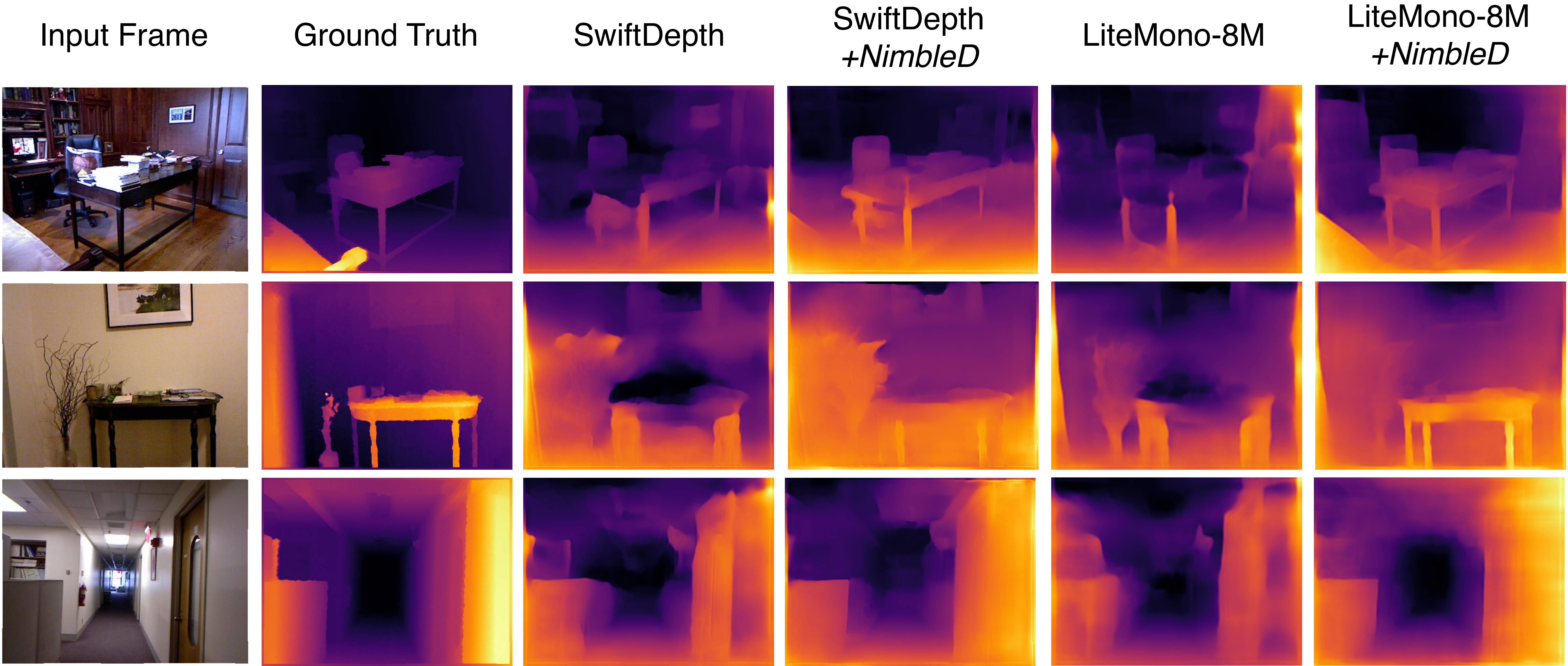}
\caption{Zero-shot qualitative results on NYUv2~\cite{nyuv2}, compared to the baseline models~\cite{swiftdepth, litemono}. NimbleD noticeably improves the generalization ability of both baseline models.}
\label{fig:nyu}
\end{figure}

\begin{figure}[t!]
\centering 
\includegraphics[width=\linewidth]{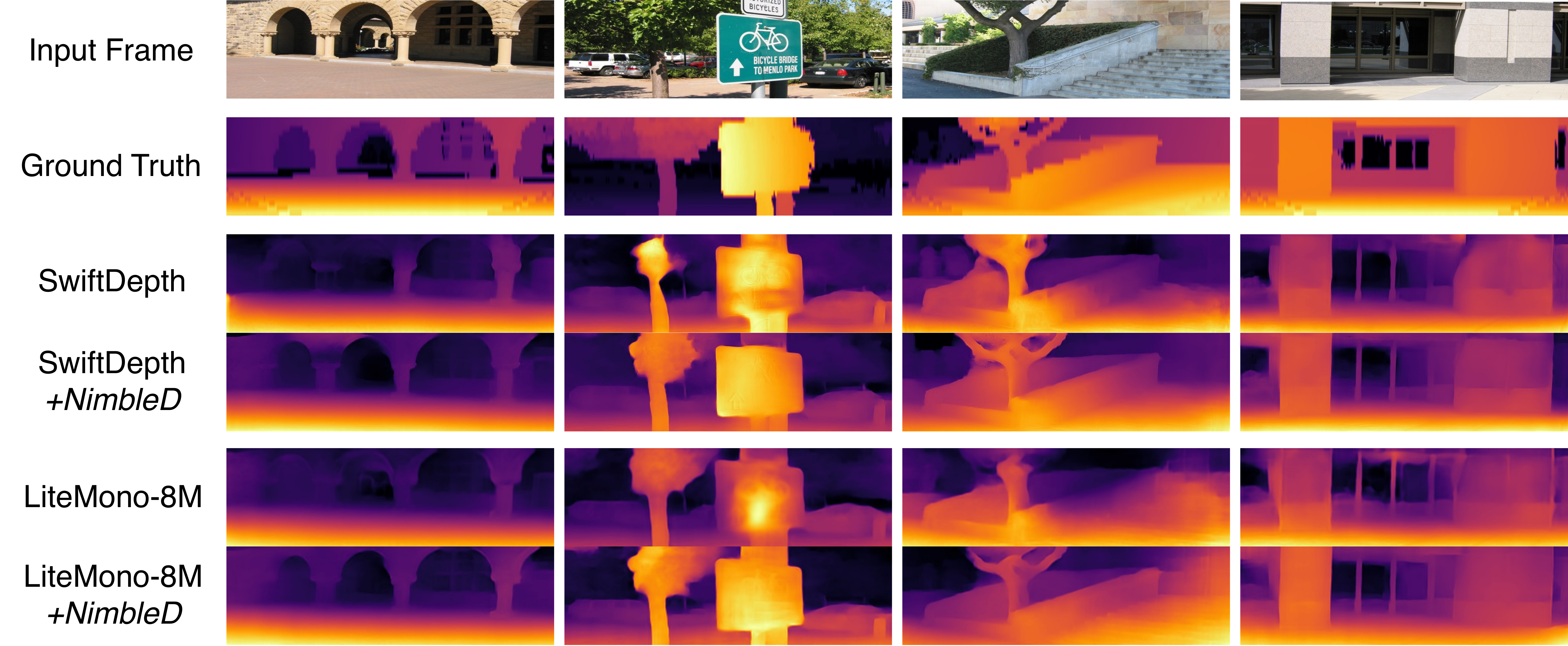}
\caption{Zero-shot qualitative results on Make3D~\cite{make3d}, compared to the baseline models~\cite{swiftdepth, litemono}. NimbleD noticeably improves the generalization ability of both baseline models.}
\label{fig:make3d}
\end{figure}

\subsubsection{Evaluation on KITTI.} 

We evaluate our method on KITTI~\cite{kitti} using Eigen split~\cite{eigen_split} with median alignment, comparing it with baseline models and leading state-of-the-art single-frame SSL MDE methods, as shown in~\cref{tab:kitti_eigen}. First, it is clear that NimbleD significantly enhances the performance of all the baseline models. The method revitalizes Monodepth2-R18~\cite{monodepth2}, while SwiftDepth~\cite{swiftdepth} and LiteMono~\cite{litemono} perform on par with FGTO~\cite{FGTO}. Meanwhile, LiteMono-S and SwiftDepth-S, which offer the best speed and accuracy trade-off on the market, are not far behind DaCCN~\cite{daccn} and GasMono~\cite{gasmono}. Moreover, LiteMono-8M demonstrates state-of-the-art performance similar to SQLdepth~\cite{sqldepth}.

We also compared our method to the teacher model, DepthAnything~\cite{depthanything}, on KITTI~\cite{kitti} Eigen-Benchmark~\cite{eigen_benchmark_split} split with improved ground truth using least squares alignment, which is the standard for evaluating supervised MDE methods, as shown in \cref{tab:kitti_eigen_benchmark_lsqr}. Since the teacher model uses a higher input image size, we also included its results with a 350$\times$350 input size for a fair comparison, which has a pixel count similar to input size of 192$\times$640. Notably, almost all of our models performed similarly to DepthAnything with an input resolution of 350$\times$350, while LiteMono-8M achieved results close to the teacher model's resolution of 518$\times$518. This demonstrates that NimbleD performs exceptionally well in offline knowledge distillation, considering the number of parameters is reduced by 38 times and the model does not use ground truth data.

The qualitative results compared with DepthAnything, Monodepth2-R50, SwiftDepth, and LiteMono-8M are illustrated in~\cref{fig:kitti}.

\subsubsection{Generalization.}

\begin{table}[t]
\centering
\caption{Zero-shot results on NYUv2~\cite{nyuv2} and Make3D~\cite{make3d} against baseline models with median alignment. The input image size is $288\times384$ for NYUv2 and $192\times640$ for Make3D. Lower values are better for all columns.}
\begin{tabular}{lcccccccc}
    \toprule
    \multirow{2.5}{*}{Method} & \multicolumn{4}{c}{NYUv2 (indoor)~\cite{nyuv2}} & \multicolumn{4}{c}{Make3D (outdoor)~\cite{make3d}} \\
    \cmidrule(lr){2-5} \cmidrule(lr){6-9}
     & AbsRel & SqRel & RMS & RMS$_{\log}$ & AbsRel & SqRel & RMS & RMS$_{\log}$ \\
    \midrule
    MD2-R18 & 0.362 & 0.665 & 1.266 & 0.390 & 0.322 & 3.587 & 7.417 & 0.163 \\ 
    \textit{$+$NimbleD} & 0.237 & 0.279 & 0.802 & 0.281 & 0.287 & 3.782 & 7.179 & 0.146 \\ 
    \midrule
    MD2-R50 & 0.326 & 0.556 & 1.139 & 0.358 & 0.322 & 3.611 & 7.326 & 0.160 \\ 
    \textit{$+$NimbleD} & 0.236 & 0.247 & 0.764 & 0.278 & 0.284 & 3.477 & 7.053 & 0.145 \\ 
    \midrule
    SwiftDepth-S & 0.317 & 0.457 & 1.070 & 0.357 & 0.330 & 3.909 & 7.547 & 0.163 \\ 
    \textit{$+$NimbleD} & \underline{0.223} & 0.240 & 0.759 & 0.267 & 0.293 & 3.834 & 7.197 & 0.146 \\ 
    \midrule
    SwiftDepth & 0.277 & 0.386 & 0.969 & 0.314 & 0.312 & \textbf{3.292} & 7.125 & 0.158 \\ 
    \textit{$+$NimbleD} & \textbf{0.204} & \textbf{0.200} & \textbf{0.689} & \textbf{0.244} & 0.282 & 3.708 & 7.100 & \underline{0.143} \\ 
    \midrule
    LiteMono-S & 0.328 & 0.537 & 1.161 & 0.362 & 0.332 & \underline{3.372} & 7.248 & 0.167 \\ 
    \textit{$+$NimbleD} & 0.236 & 0.259 & 0.782 & 0.287 & 0.300 & 4.074 & 7.382 & 0.149 \\ 
    \midrule
    LiteMono & 0.317 & 0.488 & 1.095 & 0.350 & 0.317 & 3.410 & 7.253 & 0.161 \\
    \textit{$+$NimbleD} & 0.243 & 0.310 & 0.850 & 0.295 & \underline{0.278} & 3.411 & \underline{6.979} & \underline{0.143} \\ 
    \midrule
    LiteMono-8M & 0.298 & 0.415 & 1.026 & 0.344 & 0.318 & 3.451 & 7.236 & 0.159 \\ 
    \textit{$+$NimbleD} & \textbf{0.204} & \underline{0.206} & \underline{0.699} & \underline{0.253} & \textbf{0.271} & 3.408 & \textbf{6.952} & \textbf{0.140} \\ 
    \bottomrule
\end{tabular}
\label{tab:nyuv2_make3d}
\end{table}

We evaluate the zero-shot generalization of our method on NYUv2~\cite{nyuv2} and Make3D~\cite{make3d} by comparing it to baseline models~\cite{swiftdepth, litemono}. The results are shown in~\cref{tab:nyuv2_make3d}. NimbleD significantly enhances the models' generalization ability on NYUv2 (indoor), while the improvement is less notable on Make3D (outdoor). Overall, despite major metric improvements, the results remain unsurprisingly subpar since the models are fine-tuned on KITTI~\cite{kitti}. The qualitative comparison with baseline models on NYUv2 is illustrated in~\cref{fig:nyu}, while the comparison on Make3D is depicted in~\cref{fig:make3d}.

\subsection{Ablation Study}

\begin{table}[t]
\centering
\caption{Ablation study of loss components, camera intrinsics, and pre-training on the KITTI~\cite{kitti} Eigen split~\cite{eigen_split} with median alignment. Input image size: $192 \times 640$. $Base$: baseline method which is SwiftDepth\cite{swiftdepth}. $SSL$: self-supervised learning. $PSL$: pseudo-supervised learning. $ReLU$: output activation function is ReLU instead of sigmoid. $multi$: multi-scale outputs. $photo$: photometric error. $min$: minimum reconstruction. $auto$: automasking. $sm$: smoothness loss. $K$: camera intrinsics required. $LSVP$: large-scale video pre-training. * marks increasing loss weighting $\lambda$ from 0.9 to 0.95 for the last 5 fine-tuning epochs. The highest scores are highlighted in \textbf{bold}, and the second-highest scores are \underline{underlined}.}
\begin{tabular}{cccccccccccccccc}
    \toprule
    \multirow{2.5}{*}{Loss} & \multirow{2.5}{*}{\adjustbox{valign=c}{\rotatebox{90}{ReLU}}} & \multirow{2.5}{*}{\adjustbox{valign=c}{\rotatebox{90}{multi}}} & \multirow{2.5}{*}{\adjustbox{valign=c}{\rotatebox{90}{photo}}} & \multirow{2.5}{*}{\adjustbox{valign=c}{\rotatebox{90}{min}}} & \multirow{2.5}{*}{\adjustbox{valign=c}{\rotatebox{90}{auto}}} & \multirow{2.5}{*}{\adjustbox{valign=c}{\rotatebox{90}{sm}}} & \multirow{2.5}{*}{\adjustbox{valign=c}{\rotatebox{90}{K}}} & \multirow{2.5}{*}{\adjustbox{valign=c}{\rotatebox{90}{LSVP}}} & \multicolumn{4}{c}{Lower is better} & \multicolumn{3}{c}{Higher is better} \\
    \cmidrule(lr){10-13} \cmidrule(lr){14-16}
    & & & & & & & & & AbsRel & SqRel & RMS & RMS$_{\log}$ & $\delta_1$ & $\delta_2$ & $\delta_3$ \\
    \midrule
    Base~\cite{swiftdepth} & ~ & \hspace{1pt}\checkmark\hspace{1pt} & \hspace{1pt}\checkmark\hspace{1pt} & \hspace{1pt}\checkmark\hspace{1pt} & \hspace{1pt}\checkmark\hspace{1pt} & \hspace{1pt}\checkmark\hspace{1pt} & \hspace{1pt}\checkmark\hspace{1pt} & ~ & 0.107 & 0.790 & 4.643 & 0.182 & 0.888 & 0.963 & 0.983 \\ 
    SSL & \hspace{1pt}\checkmark\hspace{1pt} & \hspace{1pt}\checkmark\hspace{1pt} & \hspace{1pt}\checkmark\hspace{1pt} & \hspace{1pt}\checkmark\hspace{1pt} & \hspace{1pt}\checkmark\hspace{1pt} & \hspace{1pt}\checkmark\hspace{1pt} & \hspace{1pt}\checkmark\hspace{1pt} & ~ & 0.108 & 0.814 & 4.624 & 0.185 & 0.886 & 0.963 & 0.982 \\ 
    SSL & \hspace{1pt}\checkmark\hspace{1pt} & \hspace{1pt}\checkmark\hspace{1pt} & \hspace{1pt}\checkmark\hspace{1pt} & \hspace{1pt}\checkmark\hspace{1pt} & \hspace{1pt}\checkmark\hspace{1pt} & ~ & \hspace{1pt}\checkmark\hspace{1pt} & ~ & 0.107 & 0.808 & 4.616 & 0.184 & 0.888 & 0.963 & 0.982 \\ 
    SSL & \hspace{1pt}\checkmark\hspace{1pt} & \hspace{1pt}\checkmark\hspace{1pt} & \hspace{1pt}\checkmark\hspace{1pt} & \hspace{1pt}\checkmark\hspace{1pt} & \hspace{1pt}\checkmark\hspace{1pt} & ~ & ~ & ~ & 0.110 & 0.878 & 4.742 & 0.187 & 0.887 & 0.962 & 0.981 \\ 
    PSL & \hspace{1pt}\checkmark\hspace{1pt} & ~ & ~ & ~ & ~ & ~ & ~ & ~ & 0.108 & 0.762 & 4.735 & 0.180 & 0.877 & 0.963 & \textbf{0.986} \\ 
    PSL & \hspace{1pt}\checkmark\hspace{1pt} & \hspace{1pt}\checkmark\hspace{1pt} & ~ & ~ & ~ & ~ & ~ & ~ & 0.108 & 0.764 & 4.708 & 0.179 & 0.878 & 0.964 & \textbf{0.986} \\ 
    SSL$+$PSL & \hspace{1pt}\checkmark\hspace{1pt} & \hspace{1pt}\checkmark\hspace{1pt} & \hspace{1pt}\checkmark\hspace{1pt} & \hspace{1pt}\checkmark\hspace{1pt} & \hspace{1pt}\checkmark\hspace{1pt} & ~ & \hspace{1pt}\checkmark\hspace{1pt} & ~ & 0.101 & 0.718 & 4.415 & 0.175 & 0.895 & \underline{0.967} & \textbf{0.986} \\ 
    SSL$+$PSL & \hspace{1pt}\checkmark\hspace{1pt} & \hspace{1pt}\checkmark\hspace{1pt} & \hspace{1pt}\checkmark\hspace{1pt} & \hspace{1pt}\checkmark\hspace{1pt} & \hspace{1pt}\checkmark\hspace{1pt} & ~ & ~ & ~ & 0.101 & \underline{0.711} & 4.435 & \underline{0.174} & 0.894 & \underline{0.967} & \textbf{0.986} \\ 
    SSL$+$PSL & \hspace{1pt}\checkmark\hspace{1pt} & \hspace{1pt}\checkmark\hspace{1pt} & \hspace{1pt}\checkmark\hspace{1pt} & \hspace{1pt}\checkmark\hspace{1pt} & ~ & ~ & ~ & ~ & 0.104 & 0.770 & 4.578 & 0.176 & 0.890 & \underline{0.967} & \underline{0.985} \\ 
    SSL$+$PSL & \hspace{1pt}\checkmark\hspace{1pt} & \hspace{1pt}\checkmark\hspace{1pt} & \hspace{1pt}\checkmark\hspace{1pt} & ~ & ~ & ~ & ~ & ~ & 0.103 & 0.731 & 4.481 & 0.175 & 0.890 & 0.966 & \underline{0.985} \\ 
    SSL$+$PSL & \hspace{1pt}\checkmark\hspace{1pt} & \hspace{1pt}\checkmark\hspace{1pt} & \hspace{1pt}\checkmark\hspace{1pt} & \hspace{1pt}\checkmark\hspace{1pt} & \hspace{1pt}\checkmark\hspace{1pt} & ~ & ~ & \hspace{1pt}\checkmark\hspace{1pt} & \underline{0.097} & \underline{0.711} & \underline{4.354} & \textbf{0.171} & \underline{0.903} & \textbf{0.969} & \textbf{0.986} \\
    SSL$+$PSL* & \hspace{1pt}\checkmark\hspace{1pt} & \hspace{1pt}\checkmark\hspace{1pt} & \hspace{1pt}\checkmark\hspace{1pt} & \hspace{1pt}\checkmark\hspace{1pt} & \hspace{1pt}\checkmark\hspace{1pt} & ~ & ~ & \hspace{1pt}\checkmark\hspace{1pt} & \textbf{0.096} & \textbf{0.697} & \textbf{4.333} & \textbf{0.171} & \textbf{0.905} & \textbf{0.969} & \textbf{0.986} \\ 
    \bottomrule\textbf{}
\end{tabular}
\label{tab:ablation_loss}
\end{table}

\begin{table}[t!]
\centering
\caption{Results on KITTI~\cite{kitti} Eigen split~\cite{eigen_split} with various values for the loss weighting parameter $\lambda$. The table presents averaged results from the last 5 epochs trained solely on KITTI for 20 epochs. The input image size is $192\times640$. The highest scores are highlighted in \textbf{bold}, and the second-highest scores are \underline{underlined}.}
\begin{tabular}{lccccccc}
    \toprule
    \multirow{2.5}{*}{$\lambda$} & \multicolumn{4}{c}{Lower is better} & \multicolumn{3}{c}{Higher is better} \\
    \cmidrule(lr){2-5} \cmidrule(lr){6-8}
     & AbsRel & SqRel & RMS & RMS$_{\log}$ & $\delta_1$ & $\delta_2$ & $\delta_3$ \\
    \midrule
    0.1 & 0.107 & 0.722 & 4.656 & 0.179 & 0.879 & 0.964 & \textbf{0.986} \\ 
    0.3 & 0.106 & 0.733 & 4.651 & 0.179 & 0.882 & 0.964 & \textbf{0.986} \\ 
    0.5 & 0.105 & 0.719 & 4.627 & 0.177 & 0.882 & 0.964 & \textbf{0.986} \\ 
    0.7 & 0.104 & 0.713 & 4.531 & \underline{0.176} & 0.887 & 0.966 & \textbf{0.986} \\ 
    0.8 & \underline{0.102} & \underline{0.703} & 4.458 & \textbf{0.175} & 0.891 & \textbf{0.967} & \textbf{0.986} \\ 
    0.85 & 0.102 & \textbf{0.702} & \underline{4.438} & \underline{0.176} & \underline{0.892} & \textbf{0.967} & \textbf{0.986} \\ 
    0.9 & \textbf{0.101} & 0.720 & \textbf{4.437} & \textbf{0.175} & \textbf{0.895} & \textbf{0.967} & \textbf{0.986} \\ 
    0.95 & 0.103 & 0.761 & 4.518 & 0.178 & \textbf{0.895} & \underline{0.966} & \underline{0.985} \\ 
    0.99 & 0.113 & 1.013 & 4.906 & 0.216 & 0.883 & 0.960 & 0.980 \\ 
    \bottomrule
\end{tabular}
\label{tab:ablation_lambda}
\end{table}

We conduct extensive experiments on the components of the loss function, the influence of training without known camera intrinsics, and the impact of large-scale pre-training. SwiftDepth~\cite{swiftdepth} is used as a baseline model. The evaluation results on KITTI~\cite{kitti} Eigen split~\cite{eigen_split} are presented in~\cref{tab:ablation_loss}. Initially, we replace the output activation function from sigmoid to ReLU in our baseline model, resulting in a minor deterioration of the results. Subsequently, we eliminate the widely-used smoothness regularization loss~\cite{monodepth2}, which does not affect the outcomes. Training without camera intrinsics notably worsens the SSL results, and we note the model's frequent convergence issues during training. We also evaluate the PSL loss calculation with both single-scale and multi-scale disparity predictions, finding no significant difference between them. To maintain consistency with the SSL loss, we opt for the multi-scale approach. Further experimentation on the combined SSL and PSL loss with known camera intrinsics reveals a significant improvement in results, indicating a synergistic effect of the combination. Interestingly, the results remain unchanged when training without camera intrinsics, highlighting the supportive role of PSL for not only improving metrics but also for making training robust in contrast to KBR~\cite{slowtv}. We then explore the necessity of minimum reconstruction and automasking~\cite{monodepth2}; however, omitting them leads to a deterioration in performance. Lastly, we apply large-scale pre-training to our model using the selected combined loss on YouTube videos, followed by fine-tuning on KITTI, which significantly enhances the evaluation metrics. Combined SSL and PSL experiments use a loss weighting parameter $\lambda$ set to 0.9, determined by extensive manual search as shown in~\cref{tab:ablation_lambda}. Interestingly, increasing $\lambda$ to 0.95 for the last 5 fine-tuning epochs slightly improves and stabilizes results.

\section{Conclusion}

In this work, we introduce NimbleD, an efficient MDE learning framework that enhances SSL using pseudo-labels generated by a large vision model. This approach allows for robust training without camera intrinsics and maintains quality, enabling us to leverage large-scale video pre-training. Consequently, NimbleD elevates the depth estimation performance of fast and lightweight models to match state-of-the-art SSL MDE methods without introducing any overhead, which is advantageous for real-time metaverse applications. While some may argue that using pseudo-supervision from larger models is inequitable, we believe that large vision models are a crucial part of the current deep learning landscape and should be fully utilized.

%
%
\bibliographystyle{splncs04}
\bibliography{main}
\end{document}